\documentclass[10pt,journal,compsoc]{IEEEtran}
\pdfoutput=1
\ifCLASSOPTIONcompsoc
  \usepackage[nocompress]{cite}
\else
  \usepackage{cite}
\fi
\ifCLASSINFOpdf
\else
\fi

\usepackage{times}
\usepackage{epsfig}
\usepackage{graphicx}
\usepackage{amsmath}
\usepackage{amssymb}
\usepackage{threeparttable}
\usepackage{slashbox}
\usepackage{epstopdf}
\usepackage{booktabs}
\usepackage[pagebackref=true,breaklinks=true,letterpaper=true,colorlinks,bookmarks=false]{hyperref}
\usepackage{multirow}
\usepackage{array}
\usepackage{color}
\usepackage{footnote}
\usepackage{caption}
\usepackage{tabularx}
\usepackage{pbox}
\usepackage{cleveref}
\usepackage{epstopdf}

\begin{document}
\title{Clustering Millions of Faces by Identity}

\author{Charles~Otto,~\IEEEmembership{Student Member,~IEEE,
	   Dayong~Wang,~\IEEEmembership{Member,~IEEE,}
	   and Anil K. Jain,~\IEEEmembership{Fellow,~IEEE}}
        }

\markboth{MSU-CSE-16-3, April 2016}%
{Shell \MakeLowercase{\textit{et al.}}: Bare Demo of IEEEtran.cls for Computer Society Journals}
\IEEEtitleabstractindextext{%
\begin{abstract}
In this work, we attempt to address the following problem: Given a large number of unlabeled face images, cluster them into the individual identities present in this data. We consider this a relevant problem in different application scenarios ranging from social media to law enforcement. In large-scale scenarios the number of  faces  in the collection can be of the order of hundreds of million, while the number of clusters can range from a few thousand to millions--leading to difficulties in terms of both run-time complexity and evaluating clustering and per-cluster quality.
An efficient and effective Rank-Order clustering algorithm is developed to achieve the desired scalability, and better clustering accuracy than other well-known algorithms such as k-means and spectral clustering. We cluster up to $123$ million face images into over $10$ million clusters, and analyze the results in terms of both external cluster quality measures (known face labels) and internal cluster quality measures (unknown face labels) and run-time. Our algorithm achieves an F-measure of $0.87$ on a benchmark unconstrained face dataset (LFW, consisting of $13$K faces), and $0.27$ on the largest dataset considered ($13$K images in LFW, plus $123$M distractor images). Additionally, we present preliminary work on video frame clustering (achieving $0.71$ F-measure when clustering all frames in the benchmark YouTube Faces dataset). A per-cluster quality measure is developed which can be used to rank individual clusters and to automatically identify a subset of good quality clusters for manual exploration.
\end{abstract}

}

\maketitle
\IEEEdisplaynontitleabstractindextext
\IEEEpeerreviewmaketitle

\IEEEraisesectionheading{\section{Introduction}\label{sec:introduction}}

\setlength{\belowcaptionskip}{-5pt}

In this work, we attempt to address the following problem: Given a large number of unlabeled face images, cluster them into the individual identities present in this data. This situation is encountered in a number of different application scenarios ranging from social media to law enforcement, where the number of  faces  in the collection can be of the order of hundreds of million.  Often, the labels attached to the face images are either missing or contain noise. The number of clusters or the unknown number of identities can range from a few thousand to  hundreds of millions, leading to difficulties in terms of both run-time and clustering quality.


Considering social media, Facebook reported that $350$ million images are uploaded per day on average\footnote{https://goo.gl/FmzROn}, and of those images, a large number may reasonably be assumed to be images of people. In social media some identity information may be provided via tagging, but in general this is incomplete and may be inaccurate. We consider grouping face images into discrete identities as one possible approach for organizing this large volume of data. 



In forensic investigations, triaging large-scale face collections is also an emerging problem. Few examples are more relevant than the Boston Marathon bombing~\cite{klontz2013case}, where tens of thousands of images and videos needed to be analyzed during a time sensitive investigation~\cite{bib:fbiva}. Other common cases that require the investigation of large media collections include identifying perpetrators and victims in child exploitation cases\footnote{http://www.nist.gov/itl/iad/ig/chexia-face.cfm}, an understanding of which individuals exist in a collection of social media (such as imagery from gang and terrorist networks), and organizing media collections from hard drives (personal computers or servers).

\begin{figure}[t]
\center
\includegraphics[width=\columnwidth]{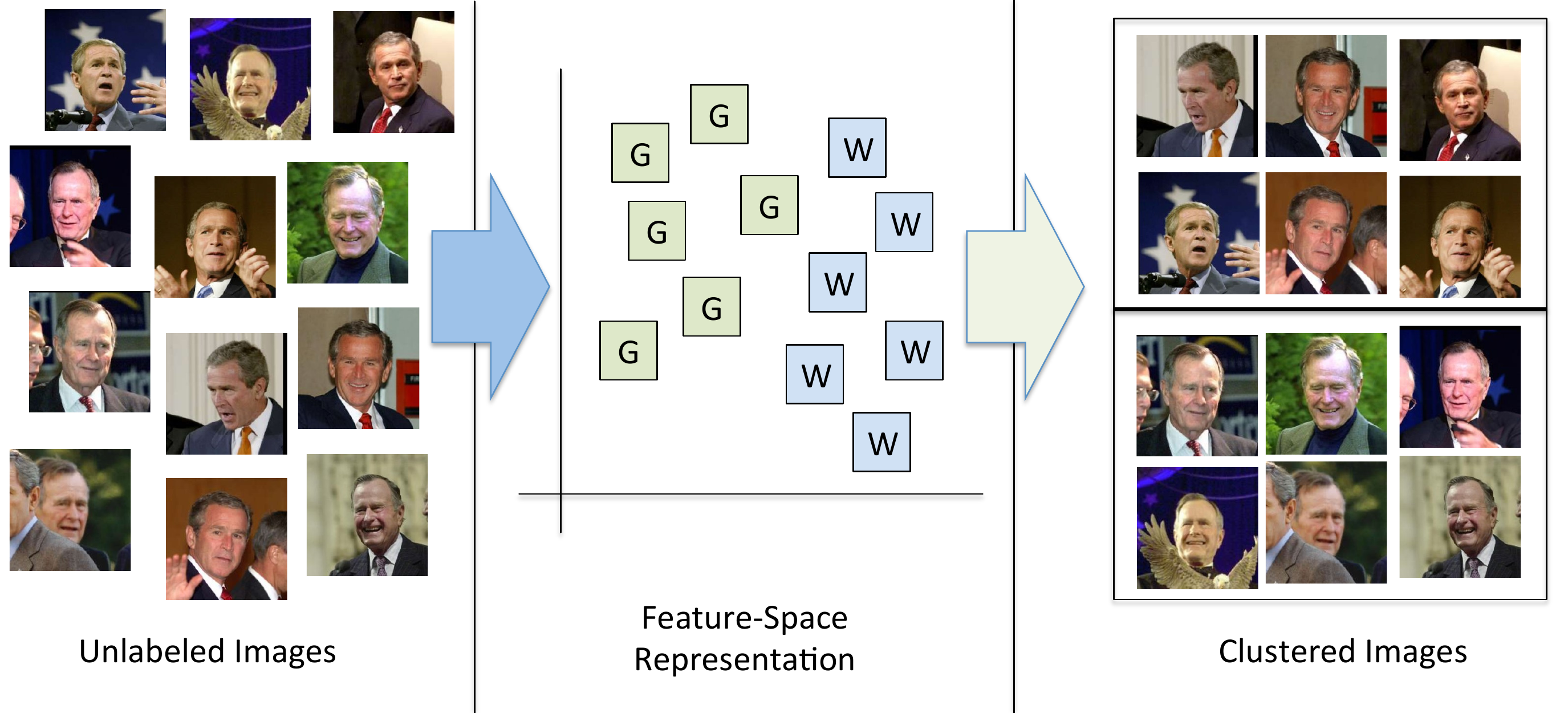}
   \caption{  Given an unlabeled set of face images acquired e.g. from social media or in the course of a forensic investigation, we propose ``clustering by identity" as a first step in exploring and understanding the dataset.  Face images here belong to two individuals: George W. Bush (W) and George H.W. Bush (G).}
\label{Fig.Problem}
\end{figure}

\label{sec:background}

\begin{table*}[t]
\centering
\captionof{table}{A summary of related studies on face clustering. }
\label{Tab.Related}

\begin{tabularx}{\textwidth}{lXlrr}
\toprule
Publication & \pbox{20cm}{Features } & \pbox{20cm}{Clustering method } & \# Face images & \# Subjects\\
\midrule
Ho et al.~\cite{ho2003clustering}   & Gradient and pixel intensity features & Spectral clustering & 1,386 & 66\\
Zhao et al.~\cite{zhao2006automatic}   & 2DHMM + contextual &Hierarchical clustering & 1,500 & 8\\
Cui et al.~\cite{cui2007easyalbum}   & LBP, clothing color + texture & Spectral & 400 & 5\\
Tian et al.~\cite{tian2007face}   & Image + contextual & Partial clustering & 1,147 & 34\\
Zhu et al.~\cite{zhu2011rank}  & Learning-based descriptor~\cite{cao2010face} & Rank-order & 1,322 & 53\\
Vidal and Favaro~\cite{vidal2014low}   & Joint subspace learning and clustering & -$^{\dag}$ & 2,432 & 38\\
Otto et al.~\cite{ottoefficient} & Component-based features, commercial face matcher & k-Means, spectral, rank-order  & 1M & 195,494 \\
\emph{Ours} & Deep features~\cite{wang2015face} & Approximate rank-order & 123M & Unknown$^{\ddag}$ \\
\bottomrule
\end{tabularx}
\vspace{-2mm}
\flushleft
$^{\dag}$ In this work a unified algorithm is used for representation and clustering

$^{\ddag}$ Due to the nature of the dataset used (face images blindly harvested from the Internet), we do not know the true number of identities, as is the case in practical scenarios.

\vspace{-4mm}

\end{table*}

In both social media, and forensic investigations we expect the unknown number of individual identities present in a dataset to be large, which is challenging from a scalability perspective since runtimes tend to be related to the number of clusters. Additionally, we expect the number of images per individual to be unbalanced (some people may appear very often, others much less frequently), which is challenging for e.g. clustering algorithms like k-means which tend to generate similar sized clusters. It can also be assumed that the quality of images in terms of pose, illumination, occlusion, etc.  being considered is relatively low, since social media images, images taken at public events etc. are not generally captured in the most favorable conditions for face recognition. Following recent progress in unconstrained face recognition, we attempt to mitigate the difficulty of the underlying face clustering  problem by using a state-of-the-art convolutional neural network based face representation~\cite{wang2015face}.

Even using a strong face representation, accuracy is not perfect on verification tasks (particularly when considering difficult data). Zhu et al.~\cite{zhu2011rank} reported success in clustering collections of personal photographs using a Rank-Order clustering method which develops a distance measure based on shared nearest-neighbors of face images being compared (since direct feature vector-to-feature vector distances may be inaccurate given the difficulty of the face recognition task). However, in addition to the problem of poor face quality, large scale face clustering tasks (on the order of $100$ million face images) are inherently difficult in terms of scalability (run-time). We develop a version of the rank-order clustering algorithm of Zhu et al.~\cite{zhu2011rank} leveraging an approximate nearest neighbor method for improved scalability, and simplifying the actual clustering procedure to achieve improved scalability and clustering accuracy.

We evaluate large-scale clustering performance by combining the well-known Labeled Faces in the Wild (LFW) dataset~\cite{LFWTech} with up to $123$M unlabeled images (downloaded from the web), and clustering the augmented dataset.  Additionally, considering that even a reasonably accurate clustering of a truly large dataset may still result in too many clusters to be manually investigated, we investigate per-cluster ``internal" quality measures (which do not require  external labels on face images)  to identify a subset of ``good" clusters (relatively compact and isolated), for manual exploration.  In addition to large-scale clustering on unconstrained still face images, we perform preliminary investigations of clustering video frames leveraging the YouTube Faces (YTF) database~\cite{wolf2011face}, clustering hundreds of thousands of video frames.

The perceived contributions of this paper include: (i) an updated clustering algorithm, improving on the method presented by Zhu et al.~\cite{zhu2011rank} using an approximate nearest neighbor method for improved scalability, which also attains better clustering accuracy, (ii) large-scale face clustering experiments using a state-of-the-art face representation learned for large scale supervised face recognition based on deep networks~\cite{wang2015face}, (iii) a preliminary investigation of the applicability of the presented face clustering method to video, and (iv) definition of a per-cluster quality measure suitable for prioritizing a subset of clusters out of millions of detected clusters.

\section{Background}

\subsection{Face Clustering}

The clustering problem, a tool for exploratory data analysis, has been well studied in pattern recognition, statistics, and machine learning literature (Jain~\cite{jain2010data} provides a survey). Less studied is the challenging problem of clustering face images, especially when both the number of images and the number of clusters are very large. An important consideration in clustering (and classifying) face images is that since there is no universally agreed upon face representation or distance metric, the clustering results depend not only on the choice of clustering algorithm, but also on the quality of the underlying face representation and metric. Table~\ref{Tab.Related} lists prior work on face clustering, with the face representation and clustering algorithm used, along with the largest dataset size employed in terms of face images, as well as number of subjects.

Ho et al.~\cite{ho2003clustering} developed variations on spectral clustering wherein the affinity matrix is computed based on (i) assuming a Lambertian object with fixed camera/object positioning, and then computing the probability that two face images are of the same object (same convex polyhedral cone in the image space), or (ii) the local gradients of the images being compared; evaluation is done on the Yale-B and PIE-$66$ datasets.

Zhao et al.~\cite{zhao2006automatic} clustered personal photograph collections. Their approach combines a variety of contextual information including time based clustering, and the probability of faces of certain people to appear together in images, with identity estimates obtained via a $2$D-HMM, and hierarchical clustering results based on body detection; a dataset of $1,500$ face images of $8$ individuals is used for evaluation.

Cui et al.~\cite{cui2007easyalbum} developed a semi-automatic tool for annotating photographs, which employs clustering as an initial method for organizing photographs. LBP features are extracted from detected faces, and color and texture features are extracted from detected bodies. Spectral clustering is performed, and the clustering results can then be manually adjusted by a human operator. Evaluation is done on a dataset consisting of 400 photographs of 5 subjects. Tian et al.~\cite{tian2007face}  further developed this approach, incorporating a probabilistic clustering model, which incorporates a ``junk" class, allowing the algorithm to discard clusters that do not have tightly distributed samples.

Zhu et al.~\cite{zhu2011rank} developed a dissimilarity measure based on the rankings of two faces being compared in each face's nearest neighbor lists (formed using a basic distance metric), and perform hierarchical clustering based on the resulting rank-order distance function. The feature representation used is the result of unsupervised learning~\cite{cao2010face}. The clustering method is evaluated on several small datasets (the largest of which contains only $1,322$ face images). Wang et al. ~\cite{wang2012scalable} primarily develop an approximate $k$-NN graph construction method; in one of their experiments they apply this method to construct the nearest neighbor lists required by~\cite{zhu2011rank}, on a dataset containing LFW and an additional $500$K unlabeled face images, and use the rank-order distance measure to produce an improved $k$-NN graph (but do not perform hard assignment of faces into clusters). 

Vidal and Favaro~\cite{vidal2014low} developed a joint subspace learning and clustering approach. It derives several subspaces from the input dataset which best capture clusters in the data. They evaluate the method on the extended Yale-B database.

In related applications, Bhattarai et al. ~\cite{bhattarai2014some} develop a semi-supervised method for organizing datasets for improved retrieval speed via hierarchical clustering. Tapaswi et al.~\cite{tapaswi2014total} address organization of video frames, performing both within video and cross-video clustering, incorporating constraints from face tracking and common video editing patterns. Schroff et al. ~\cite{schroff2015facenet} give some qualitative results of clustering personal photos using a deep learning based face representation.





Some experimental work in face clustering has considered hundreds of thousands of images, while some general object clustering tasks have used datasets on the order of billions of images~\cite{liu2007clustering}. In cases where the true number of clusters is known a priori, that number is typically orders of magnitude lower than the number of images. In general, the evaluation methods used to determine how well clustering algorithms perform (when true labels are available) are split. In some cases the clustering accuracy is used~\cite{ho2003clustering}, in others precision/recall~\cite{zhao2006automatic}, and in still others normalized mutual information is employed~\cite{zhu2011rank}. 

\subsection{General Image Clustering}
For clustering images in general, rather than faces in particular, Liu et al.~\cite{liu2007clustering} (i) extracted Haar wavelet features from images, (ii) applied a distributed algorithm consisting of an approximate nearest neighbor step, (iii) generated an initial set of clusters by applying a distance threshold to the nearest neighbor lists, and (iv) applied a union-find algorithm to get a final set of clusters. Clustering was performed on approximately $1.5$ billion unlabeled images, along with an evaluation on $3,385$ labeled images. The main goal of the procedure was to group images into sets of near duplicates, but the total number of such sets in the $1.5$ billion image dataset was unknown.
	
Gong et al.~\cite{gong2015web} develop a version of k-means clustering which is suitable for handling large datasets by encoding their feature vectors to binary vectors, and then using an indexing scheme to support constant time lookup of cluster centers for the assignment step of k-means. They apply their binary k-means algorithm to a subset of the ImageNet dataset, containing $1.2$ million general object images in $1,000$ classes.

Foo et al.~\cite{foo2007clustering} consider a related problem, the detection of near-duplicate images in large datasets. In this case, rather than grouping images of people by identity, the goal is to identify near-duplicate images, which may be the result of various image processing operations, such as cropping, rotation, colorspace conversion, etc. Their image representation consists of applying a visual words approach to local PCA-SIFT descriptors, indexed with a Locality Sensitive Hashing (LSH) scheme. The clustering method used is a union-find algorithm. Evaluation was performed by generating a synthetic set of near duplicate images, and performing clustering in the presence of a separate noise set; the largest dataset used contained $300,000$ images.

\subsection{Approximate Nearest Neighbor Methods}

A common problem in some of the well-known clustering methods is finding nearest neighbor sets for all $n$ samples in a dataset. Naively, the runtime is $O(n^2)$, which is a problem for large $n$. This can be considered an instance of the $k$-NN graph construction problem, or alternatively it can be considered a set of $n$ approximate nearest neighbor searches. For both of these cases, approximation methods are available in the literature.

\subsubsection{$k$-nn Graph Construction}
One approximation method for computing the full $k$-NN graph is given by Chen et al.~\cite{chen2009fast}. The algorithm is a procedure based on recursive subdivision of the feature space via Lanczos bisection. We use a parallelized version of this algorithm, presented in ~\cite{ottoefficient}, which branches at each recursive subdivision, handling both halves in separate threads.


This algorithm achieves improved runtime over the brute-force method by skipping some sets of comparisons (the portion of comparisons at each split between samples in opposite partitions, not included in the overlap set), and as such the runtime is a function in the degree of overlap chosen.

\begin{figure}[t]
  \centering
    \includegraphics[width=\columnwidth]{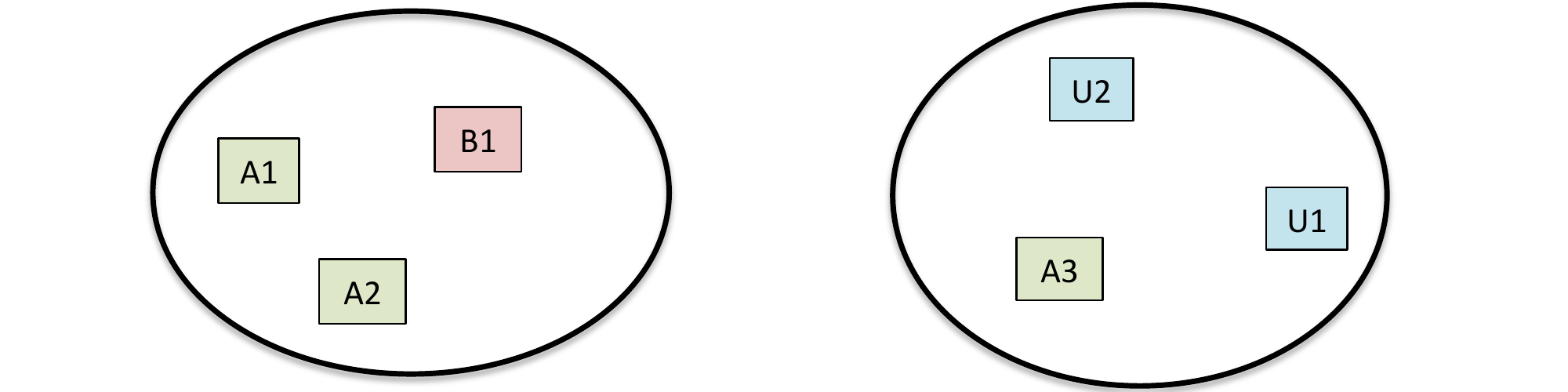}
  \caption{Diagram of a possible clustering configuration, used to illustrate evaluation metrics. Six samples are partitioned into $2$ clusters; A$1$, A$2$, and A$3$ are labeled with the same identity, sample B$1$ is labeled with a different identity, and samples U$1$ and U$2$ are unlabeled.}
  \label{fig:prec_recall_examples}
\end{figure}

\begin{figure*}[t]
  \centering
    \includegraphics[width=\linewidth]{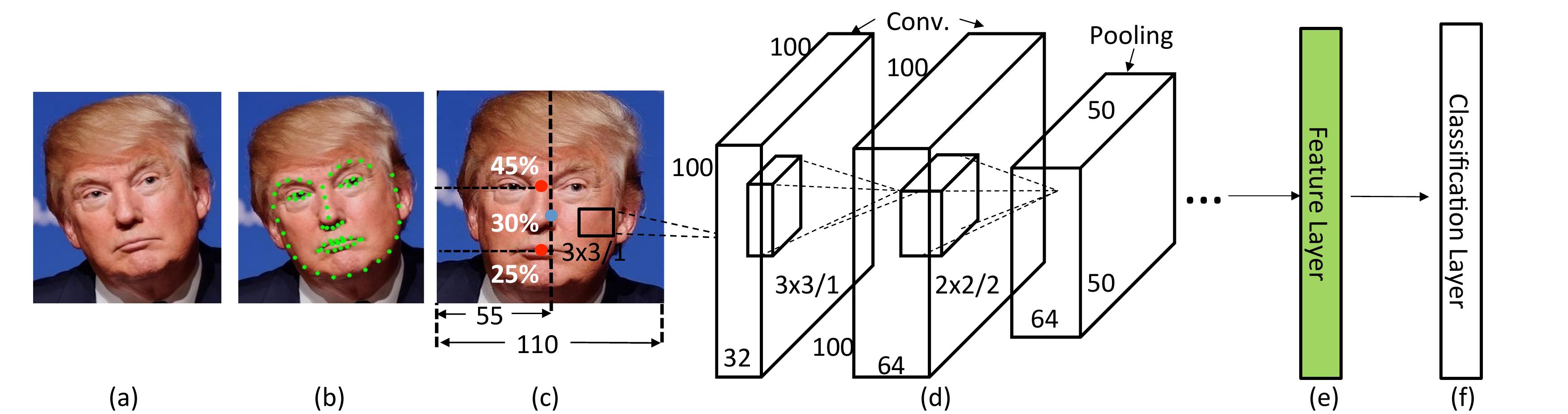}
  \caption{Face representation. An RGB image is input (a), keypoints are detected (b), the image is normalized following the procedure described in~\cite{wang2015face} (c), the normalized image is input to a convolutional neural network (d), and the $320$-dimensional output of the final average-pooling layer is used as the face representation (e). An N-way softmax classification layer (f) is used during training only.}\label{fig:architecture}
\end{figure*}

\subsubsection{Randomized k-d Tree}
In addition to $k$-NN graph construction, we may consider building nearest neighbor lists for the entire dataset as $n$ discrete nearest neighbor search problems, and improve the total runtime by employing an approximate nearest neighbor search method. Among various approximate nearest neighbor algorithms, one classic family is partitioning tree-based approaches. They follow the classic k-d tree algorithm which develops an index that subdivides the feature space by selecting a subset of features to split the data on. The randomized k-d tree algorithm~\cite{silpa2008optimised} improves efficiency by building multiple randomized k-d tree indices, then searches those indices in parallel. 

\subsection{Clustering Evaluation}

In evaluating clustering performance, since we use a pre-defined definition of ``correct" clustering (clustering by identity), we can evaluate accuracy in terms of clusters corresponding to known identity labels. \emph{External} measures for evaluating clustering quality rely on identity labels; we will use pairwise precision/recall since it can be computed efficiently.  Run time is also an important evaluation metric.

\emph{Pairwise precision} is defined as the fraction of pairs of samples within a cluster (considering all possible pairs) which are of the same class (have the same identity), over the total number of same-cluster pairs within the dataset. In Figure~\ref{fig:prec_recall_examples}, (A$1$, A$2$) is a matching pair, and (A$1$, B$1$) and (A$2$, B$1$) are mismatched pairs.

\emph{Pairwise recall} is defined as the fraction of pairs of samples within a class (considering all possible pairs) which are placed in the same cluster, over the total number of same-class pairs in the dataset. In Figure~\ref{fig:prec_recall_examples} (A$1$, A$2$) is a same class pair in the same cluster, while (A$1$, A$3$) and (A$2$, A$3$) are same-class pairs in different clusters. 

These measures capture two types of error, a clustering which places all samples as individual clusters will have high precision, but low recall, while a clustering which places all samples in the same cluster will have high recall, but low precision. The two numbers can be summarized using F-measure, defined as $F = 2\times (Precision\times Recall) / (Precision+Recall)$.

We extend these measure to handle partially labeled data, as encountered in large-scale clustering problems, by simply omitting the unlabeled data from evaluation, to the extent possible. In our experiments, partially labeled data occurs when we mix LFW face images (with known labels) against a large collection of faces downloaded from the web with unknown labels. 

We define modified pairwise recall by simply not counting whether or not unlabeled identities are grouped together. For precision, we consider labeled-unlabeled pairs (e.g. (A$3$, U$1$) and (A$3$, U$2$) in Figure~\ref{fig:prec_recall_examples}) mismatches, and omit unlabeled-unlabeled pairs (i.e. (U$1$, U$2$) ) from the calculation. So, rather than considering all possible pairs in a given cluster, we omit any unlabeled-unlabeled pairs from the total. In the right cluster in Figure~\ref{fig:prec_recall_examples}, we would only use pairs (A$3$, U$1$) and (A$3$, U$2$) to calculate the modified precision. The modified precision in Figure~\ref{fig:prec_recall_examples} is then $1 / 5$  (only the A$1$-A$2$ pair is correct, the U$1$-U$2$ pair is not counted), the modified recall is $1 / 3$, only class A has more than one sample (and is labeled), and of the class A pairs, only A$1$ and A$2$ are in the same cluster.


\section{Proposed Face Clustering Approach}
\subsection{Face Representation}
Since we are clustering faces captured under unconstrained conditions, we leverage a deep convolutional neural network for our face representation following the success of such methods by various researchers on the LFW benchmark\footnote{\url{http://vis-www.cs.umass.edu/lfw/results.html}}. Many deep learning approaches have been successfully applied to the LFW benchmark; however, most leverage private training sets. In our case, we use the architecture described in~\cite{wang2015face} and train the network directly on aligned face images from the publicly available CASIA-webface dataset~\cite{yi2014learning}. Results on both the LFW and IJB-A~\cite{klare2015pushing} benchmarks, and under larger-scale face retrieval scenarios, using this trained network, were shown to be reasonably competitive in~\cite{wang2015face}, compared to the best approaches on LFW, particularly considering the different scales of training data involved.

The feature extraction process is outlined in  Figure~\ref{fig:architecture}. Given an input image, $68$ facial landmarks are detected using the DLIB implementation of Kazemi and Sullivan's~\cite{kazemi2014one} ensemble of regression trees method. Image normalization is performed based on the detected keypoints, in particular in-plane rotation is corrected based on the angle between the eyes, the eye line is placed at $45$\% of image height from the top of the image, the mouth line is placed at $25$\% of image height from the bottom of the image, the midpoint of all detected points is centered in the x dimension, the aligned image is scaled to $110 \times 110$, and the center $100 \times 100$ region is the final normalized image.

The normalized image is passed as input to a convolutional neural network following a very deep architecture~\cite{simonyan2014very}, with a total of $10$ convolution layers, and small $(3 \times 3)$ filters. The architecture consists of pairs of convolutional layers followed by max-pooling layers, repeated $4$ times, then a final $2$ convolutional layers followed by an average pooling layer, with ReLU neurons following all convolutional layers, except for the last one. The $320$-dimensional output of the final average pooling layer is used as our feature vector, and during training is fed into a fully connected layer (regularized via dropout), followed by a softmax loss. Only the $320$-dimensional output of the average-pooling layer is used in our clustering experiments.

\begin{figure*}[t]
        \includegraphics[width=\textwidth]{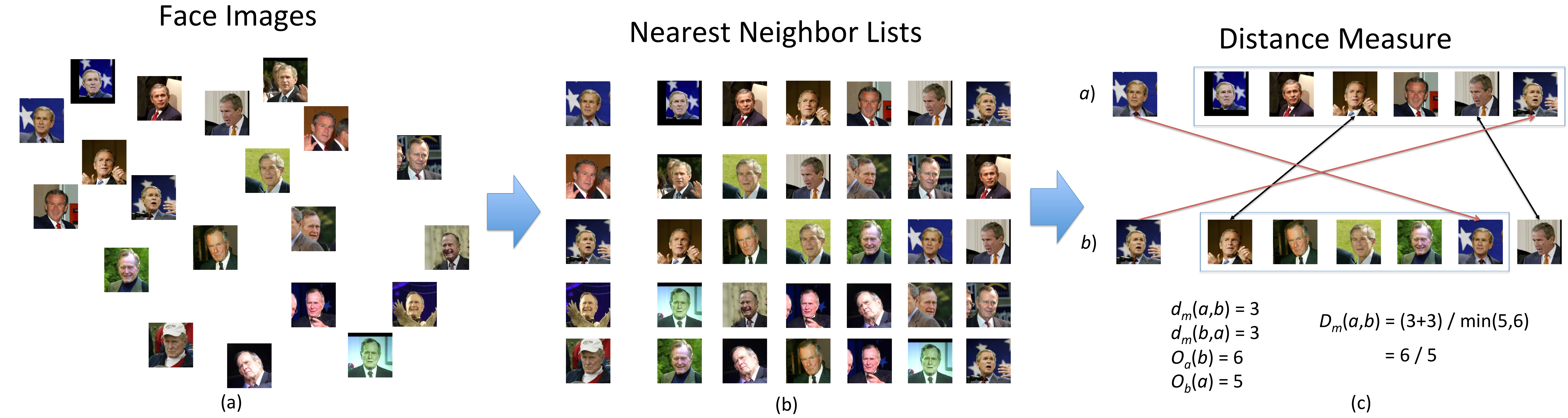}
    \caption{Approximate Rank-Order clustering. Given a set of unlabeled face images (a), nearest neighbor lists are computed for each image (b); nearest neighbor lists are then used to compute distances between faces (c). (b) shows the nearest neighbor lists of only five faces in (a). $d_m(a,b)$ (Eq. ~\ref{asym_mod}) is the asymmetric distance between faces $a$ and $b$ whereas $D_m(a,b)$ (Eq. ~\ref{eq_d_mr}) is the symmetric distance between faces $a$ and $b$.}
    \label{fig:rank_order_overview}
\end{figure*}

The network is trained using $404,992$ face images of $10,533$ subjects from the CASIA-webface dataset (the images for which face alignment was performed successfully), in minibatch stochastic gradient descent. The loss layer used for training is a single softmax loss function. The weight decay of all layers is set to $5 \times 10^{-4}$, and the learning rate for stochastic gradient descent (SGD) is initialized to $10^{-2}$, and gradually reduced to $10^{-5}$. The network is implemented using the cuda-convnet2 library\footnote{\url{https://code.google.com/p/cuda-convnet2/}}. 

\subsection{Clustering Method}
A large number of clustering methods have been proposed in the literature based on squared-error, mixture models, nearest neighbor and graph-theoretic approaches~\cite{jain2010data}. Based on evaluation of different approaches for face clustering in~\cite{ottoefficient}, we leverage an approximate version of the rank-order clustering algorithm proposed by Zhu et al.~\cite{zhu2011rank}. We present the original algorithm in detail, then our modified version.
\subsubsection{Rank-Order Clustering}
The rank-order clustering algorithm proposed by Zhu et al.~\cite{zhu2011rank}, similar to the method of Gowda and Krishna~\cite{gowda1978agglomerative}, is a form of agglomerative hierarchical clustering, using a nearest neighbor based distance measure. The overall procedure for agglomerative hierarchical clustering, given some distance metric, is to initialize all samples to be separate clusters and then iteratively merge the two closest clusters together. This requires defining a cluster-to-cluster distance metric. In the algorithm, the distance between two clusters is considered to be the minimum distance between any two samples in the clusters.

The first distance metric used in Rank-Order clustering
is given by:
\begin{equation}
d(a,b) = \sum_{i=1}^{O_a(b)}O_b(f_a(i)),
\end{equation}
\noindent where $f_a(i)$ is the $i$-th face in the neighbor list of a, and $O_b(f_a(i))$ gives the rank of face $f_a(i)$ in face b's neighbor list. This asymmetric distance function is then used to define a symmetric distance between two faces, a and b, as:

\begin{equation}
D(a,b) = \frac{d(a,b) + d(b,a) }{\textrm{min}(O_a(b), O_b(a))}.
\end{equation}

\begin{figure*}[t]
  \centering
\begin{center}
    \includegraphics[width=0.9\linewidth]{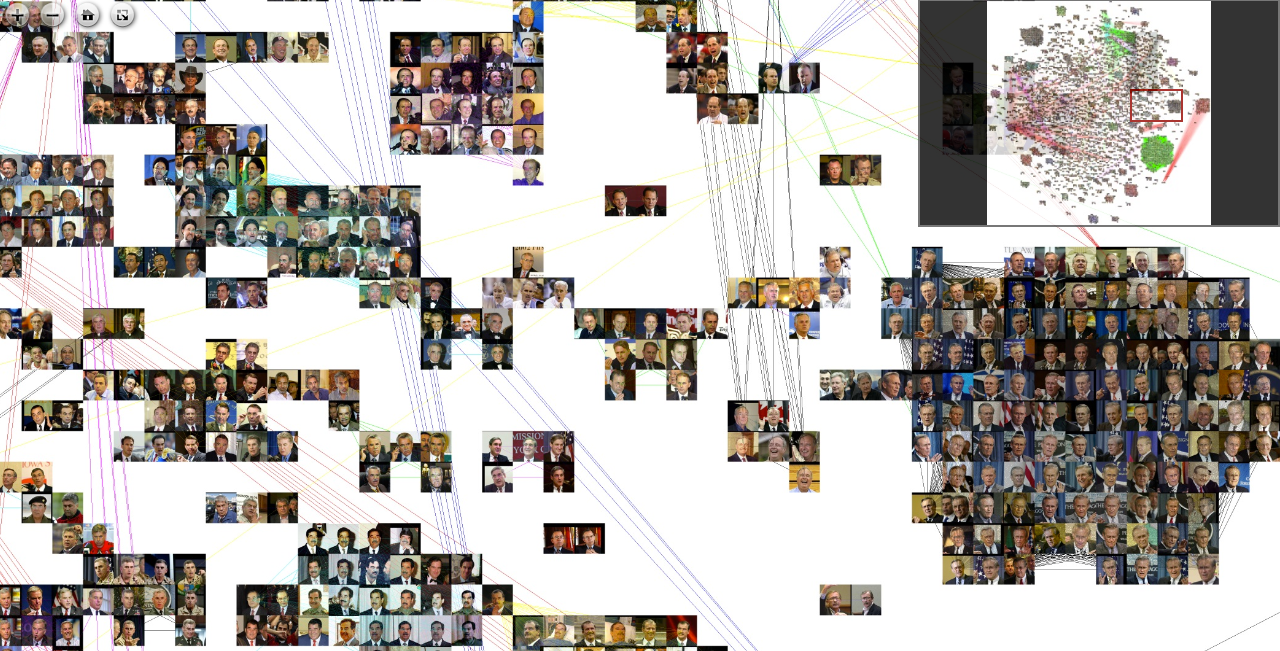}
\end{center}
  \caption{Two-dimensional t-SNE~\cite{van2008visualizing} embedding of 320-dimensional deep features for the LFW dataset, including only clusters from the proposed clustering with two or more images. Lines are drawn between all same-cluster faces.}\label{fig:embed_figure}
\end{figure*}

The symmetric rank order distance function gives low values if the two faces are close to each-other (face $a$ ranks high in face $b$'s neighbor list, and face $b$ ranks high in face $a$'s neighbor list), and have neighbors in common (high ranking neighbors of face $b$ also rank highly in face $a$'s neighbor list). After distances are computed, clustering is performed by initializing every face image to its own cluster, then computing the symmetric distances between each cluster, and merging any clusters with distance below a threshold. Then, nearest neighbor lists for any newly merged clusters are merged, and distances between the remaining clusters are computed again iteratively, until no further clusters can be merged. In this case, rather than specifying the desired number of clusters C, a distance threshold is specified; it is the threshold that determines the specific number of clusters for a particular dataset being clustered, and effective threshold values are empirically determined. We use our own implementation of this algorithm.

In terms of run-time, computing the full nearest neighbor lists for each sample incurs an $O(n^2)$ cost. Additionally, the actual clustering step used here is iterative, with cost per iteration proportional to the current number of clusters squared (with number of clusters starting at $n$ and decreasing across iterations), so both the nearest neighbor computation, and the clustering step itself are costly with increasing dataset size.

\subsubsection{Proposed Approximate Rank-Order Clustering}

The Rank-Order clustering method has an obvious scalability problem in that it requires computing nearest neighbor lists for every sample in the dataset, which has an $O(n^2)$ cost if computed directly. Although various approximation methods exist for computing nearest neighbors, they are typically only able to compute a short list of the top $k$ nearest neighbors efficiently, rather than exhaustively ranking the dataset. We use the FLANN library implementation of the randomized k-d tree algorithm~\cite{flann_pami_2014} to compute a short list of nearest neighbors. 

Applying approximation methods for faster nearest neighbor computation then requires some modification of the original Rank-Order clustering algorithm. In particular, rather than considering all neighbors in the summation equation (1), we sum up to at most the top $k$ neighbors (under the assumption that cluster formation relies on local neighborhoods).

Further, we note that if only a short list of the top-$k$ neighbors is considered, the presence or absence of a particular example on the short list may be more significant than the sample's numerical rank. As such, we consider a distance measure based on directly summing the presence/absence of shared nearest neighbors, rather than the ranks, resulting in the following distance function:
\begin{equation} \label{asym_mod}
d_m(a,b) = \sum_{i=1}^{\textrm{min}(O_a(b),k)}I_b(O_b(f_a(i)), k),
\end{equation}
\noindent where $I_b(x,k)$ is an indicator function with a value of 0 if face $x$ is in face $b$'s top $k$ nearest neighbors, and 1 otherwise. In practice, we find that this modification leads to better clustering accuracy compared to summing the ranks directly, as in the original formulation. Effectively, this distance function implies that the presence or absence of shared neighbors towards the top of the nearest neighbor list (say within the top-200 ranks) is important, while the numerical values of the ranks themselves are not.

The normalization procedure employed in the original algorithm (only summing up to the rank of the other sample being compared, and dividing by $\textrm{min}(O_a(b), O_b(a))$) is still effective, and contributes to more accurate clustering results even with this modification to the original algorithm. The combined modified distance measure is defined as:
\begin{equation}\label{eq_d_mr}
D_m(a,b) = \frac{d_m(a,b) + d_m(b,a) }{\textrm{min}(O_a(b), O_b(a))}.
\end{equation}
Additionally, to improve the runtime of the clustering step itself, we 1) only compute distances between samples which share a nearest neighbor, and 2) only perform one round of merges of individual faces into clusters. This means that compared to the original algorithm which has a runtime of $C^2$ per clustering iteration, we only perform one iteration of clustering, and additionally only check for merges on a subset of all possible pairs (since we consider the $200$ nearest neighbors for each sample), meaning that the final runtime of the clustering step (assuming pre-computed nearest neighbors) is $O(n)$. 

The final clustering procedure we employ is then:
\begin{enumerate}
  \item Extract deep features for every face in the dataset
  \item Compute a set of the top-$k$ nearest neighbors for each face in the dataset
  \item Compute pairwise distances between each face and its top-$k$ nearest neighbor lists following equation \ref{eq_d_mr}
  \item Transitively merge all pairs of faces with distances below a threshold
\end{enumerate}

Selecting a threshold to determine the number of clusters, $C$ in a given dataset is one of the perennial difficult issues in data clustering. In practical applications we cannot assume that the true number of clusters will be known a priori, therefore in the absence of a robust procedure for determining the true number of clusters, we simply evaluate our algorithm at several effective values of $C$ and report the best results attained in our experiments.

\begin{figure*}[t]
	\begin{center}
        \includegraphics[width=0.9\textwidth]{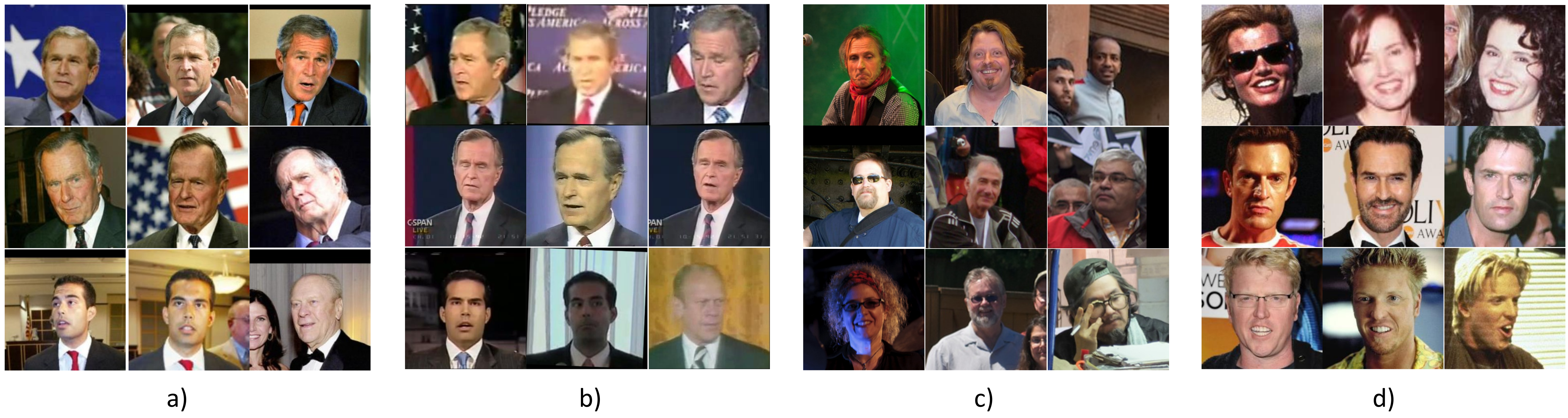}
	\end{center}
    \caption{Example face images from the a) LFW, b) Youtube Faces, c) Webfaces, and d) CASIA-webface datasets.}
    \label{fig:datasets}
\end{figure*}

\subsection{Per-Cluster Quality Evaluation}

Our overall goal is to facilitate the investigation of very large collections of unlabeled face images. We have proposed clustering face images by identity as a first approach, but for very large datasets even clustering by identity may leave too many clusters for manual exploration. We attempt to address this issue by using internal cluster validity measures to identify a subset of ``good" individual clusters, suitable for manual investigation.


In practical applications where the dataset is completely unlabeled, evaluating clustering according to external labels is not possible. But, there is a body of work in the literature on different internal cluster quality measures ~\cite{jainClusterBook} which attempt to characterize cluster quality without the use of labels. These measures can typically be understood as measures of either \emph{compactness} (how well the cluster members are grouped together in terms of pairwise similarity), or \emph{isolation} (how well different clusters are separated from each other in terms of inter-cluster similarity). Additionally, we can make a distinction between evaluating the overall quality of a given clustering of a dataset, and evaluating the quality of individual clusters in a particular clustering; we will use per-cluster quality measures as a means of ranking individual clusters. 


When dealing with very large datasets, one fundamental concern is run-time. It is generally infeasible to compute distances between all samples in the dataset, and additionally infeasible to compute distances between all clusters in cases where both the dataset and number of clusters present in the dataset are large. In this case, we are pre-computing a $k$-nearest neighbor graph, so it is natural to consider graph-based quality measures, and alleviate computational concerns by using the pre-computed graph.

Coverage~\cite{brandes2003experiments} is defined as the fraction of intra-cluster edges present out of the complete set of edges in the graph. We modify this for use as a per-cluster quality measure by just considering nodes in the current cluster, i.e. define per-cluster coverage as the fraction of edges out-bound from nodes in the current cluster which link to other nodes in the cluster. Modularization quality~\cite{mancoridis1998using} is defined as the difference between an inter-cluster connectivity measure (the fraction of edges present between nodes in a cluster out of possible edges in a complete graph of those nodes), and an intra-cluster connectivity measure (average fraction of cross-cluster edges present out of possible edges between each pair of clusters in a complete graph).

These graph-based measures are formulated solely in terms of the presence or absence of edges between certain vertices. But we also find motivation to look at some simple distance based measures by looking at Figure~\ref{fig:embed_figure} which shows a $2$-dimensional t-SNE~\cite{van2008visualizing} embedding of the original $320$-dimensional feature space for non-singleton clusters generated from the full LFW dataset. In this visualization, lines are drawn between all images placed in the same clusters. One thing which is apparent is that some cluster assignment errors cover a large distance in the embedding, and may indicate that these errors occur over a large distance in the original feature space. We will therefore also consider simple compactness and isolation measures based on the distances between edges present in the $k$-NN graph, primarily the average distance of samples to other samples in the same cluster in their nearest neighbor lists (average intra-cluster distance), and the average distance of samples in a cluster to samples outside that cluster in their nearest neighbor lists (average inter-cluster distance).

\section{Datasets}
Our clustering experiments use several unconstrained face datasets, the CASIA-webface face dataset~\cite{yi2014learning} for training the deep network feature representation, the Labeled Faces in the Wild (LFW)~\cite{LFWTech} and YouTubeFaces (YTF)~\cite{wolf2011face} datasets for clustering evaluation, and a collection of 123M unlabeled web face images used to augment the labeled datasets for larger-scale clustering evaluation. Example face images from each dataset are shown in Figure~\ref{fig:datasets}.

\begin{itemize}

\item \textbf{LFW}~\cite{LFWTech}: LFW contains $13,233$ face images of $5,749$ individuals; of those $5,749$ individuals, $4,069$ have only one face image each. The dataset was constructed by searching for images of celebrities and public figures, and retaining only images for which an automatically detectable face was present.

\item \textbf{YTF}~\cite{wolf2011face}: Similar in spirit to LFW, the YouTube Faces (YTF) dataset consists of videos of celebrities and public figures harvested from the Internet. The dataset contains $1,595$ subjects (which are a subset of the subjects in LFW), in $3,425$ videos, consisting of a total of $621,126$ individual frames. Labels are provided for the subject of interest for every frame of video where a face could be detected. In our experiments we use the pre-cropped frames, to avoid confusion between the primary subject in each video, and any unlabeled individuals that may be in a given frame.

\item \textbf{Webfaces}: To evaluate our clustering method on larger scale datasets, a cooperating research group used a crawler to automatically download a total of $123,654,141$ web images. Similar to LFW, these images were filtered to contain faces detectable by an automatic face detector, in particular the DLIB face detector.\footnote{In some cases, the detected faces are in fact false positive detections (e.g. non-human faces (such as cartoons), or non-face objects). We estimate approximately $2$\% of the total detections may be false positives, based on a manual examination of a random sample of $10,000$ detections from the full dataset. We did not delete the identified non-human faces from the dataset.}

\item \textbf{CASIA-webface}~\cite{yi2014learning}: The CASIA-webface dataset contains $494,414$ images of $10,575$ subjects (mostly celebrities); however, we are unable to localize faces in some of the images, and so use a subset of $404,992$ face images of $10,533$ subjects to train our network~\cite{wang2015face}.

\begin{figure*}[t]
\begin{center}
        \includegraphics[width=0.9\textwidth,keepaspectratio]{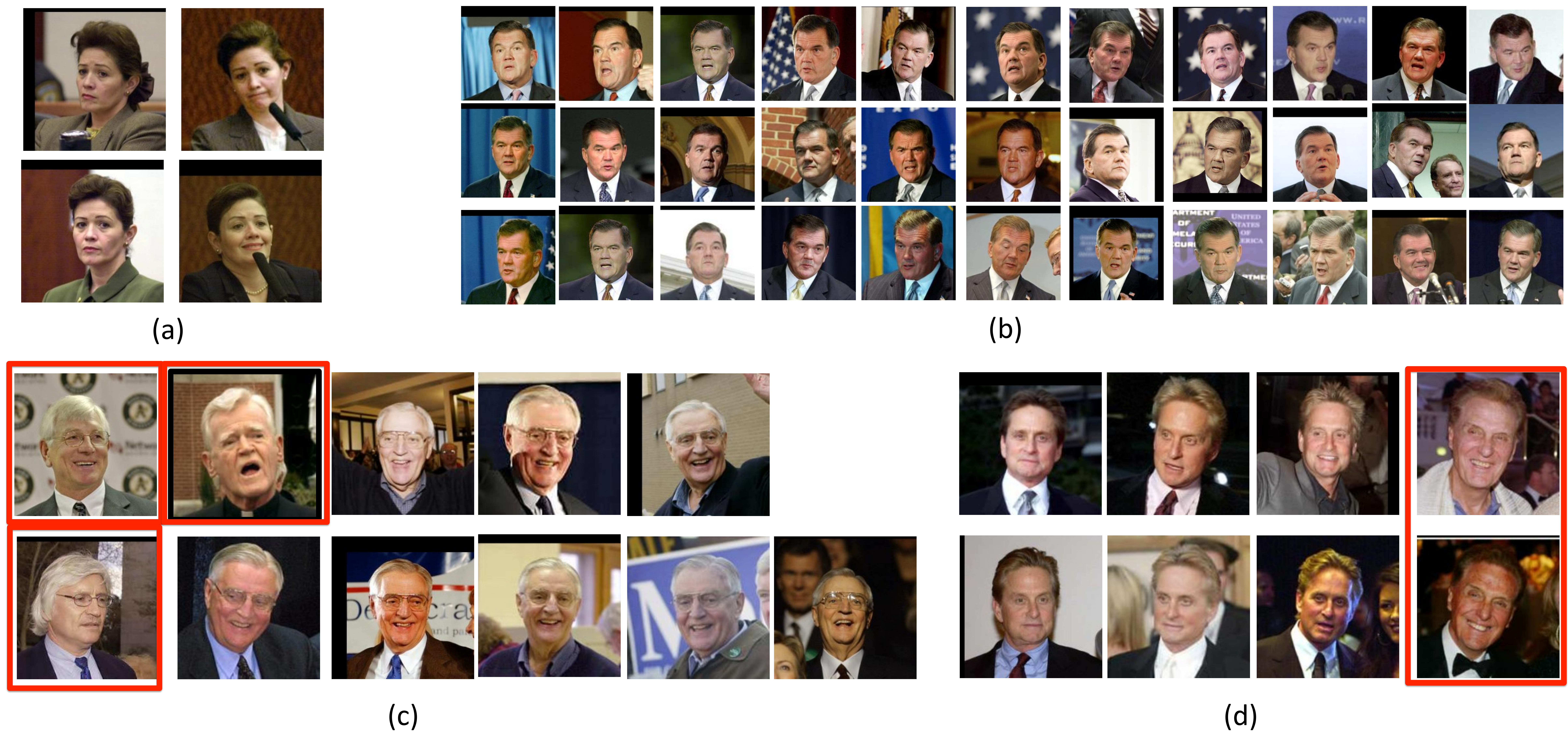}
\end{center}
    \caption{Examples of ``pure" (single individual) clusters (a, b), and ``impure" (multiple individuals) clusters (c,d) generated by the proposed Approximate Rank-Order clustering on the entire LFW dataset. Faces not belonging to the majority identity in each cluster are outlined in red.}
    \label{fig:clusters_just_lfw}
\end{figure*}


\end{itemize}

\section{Experiments}
In this section, we will present our overall evaluation of large-scale face clustering, in several steps. First, we will evaluate various clustering algorithms on a small dataset (the entire LFW dataset), evaluate the nearest neighbor approximation used, then carry out large-scale face clustering experiments (involving up to a $123$ million face dataset), and finally present some preliminary work on video clustering.

\subsection{Clustering Algorithm Evaluation}
Before investigating performance on large-scale datasets, we will attempt to cluster the entire LFW dataset by identity. One issue is that the distribution of images per subject is quite imbalanced in LFW (indeed, the majority of subjects have only a single image, accounting for approximately a third of all images in the dataset). Although we could conceivably construct a subset of LFW with more ``balanced" clusters, in practice for the application domains of large-scale clustering (analyzing social media imagery, and forensic applications), there is no basis to assume that the number of images per subject is well balanced. In the absence of prior knowledge about the expected distribution of images per subject in practical applications, we cluster the entire LFW dataset.

As a baseline, we will consider k-means clustering, since (i) it is perhaps the most well-known clustering algorithm, (ii) has only a few parameters for tuning, (iii) is one of the most efficient, and (iv) large-scale clustering methods are often approximations of k-means clustering with improved scalability.  We use the MATLAB r2015a implementation of the k-means algorithm, with the euclidean distance metric. We additionally use spectral clustering~\cite{von2007tutorial}, which approaches the problem from a graph theory perspective, as a baseline. We induce a graph structure in the adjacency matrix by keeping the top $200$ neighbors non-zero (this value seems effective in capturing local structures), and again use Euclidean distance. The number of clusters $C$ must be specified. We use a MATLAB implementation of spectral clustering\footnote{\url{http://www.mathworks.com/matlabcentral/fileexchange/34412-fast-and-efficient-spectral-clustering/content/files/SpectralClustering.m}}. Additionally, we use our implementation of the original rank-order clustering algorithm as a baseline.

\begin{table}
\centering

\caption{Clustering results on the complete LFW dataset. Times are given as HH:MM:SS, measured using 20 cores of an Intel Xeon CPU clocked at 2.5 GHz. The proposed algorithm (last row) has the highest clustering accuracy (F-measure) and the shortest run-time.}
\begin{tabular}{ l|r|r|r }
\toprule
 Clustering Algorithm & \# Clusters & F-measure & Run-Time \\ \midrule
  k-Means & 100 & 0.36 & 00:00:16 \\
  k-Means & 6,508& 0.07 & 04:58:49 \\
  Spectral & 200 & 0.20  & 00:11:18 \\
  Rank-Order & 6,591 & 0.80 & 00:00:18 \\
  Approx. Rank-Order (proposed) & 6,508 & 0.87 & 00:00:08 \\ \bottomrule
\end{tabular}

\label{tab:just_lfw}
\end{table}

\begin{figure*}[t]
\begin{center}
        \includegraphics[width=0.9\textwidth]{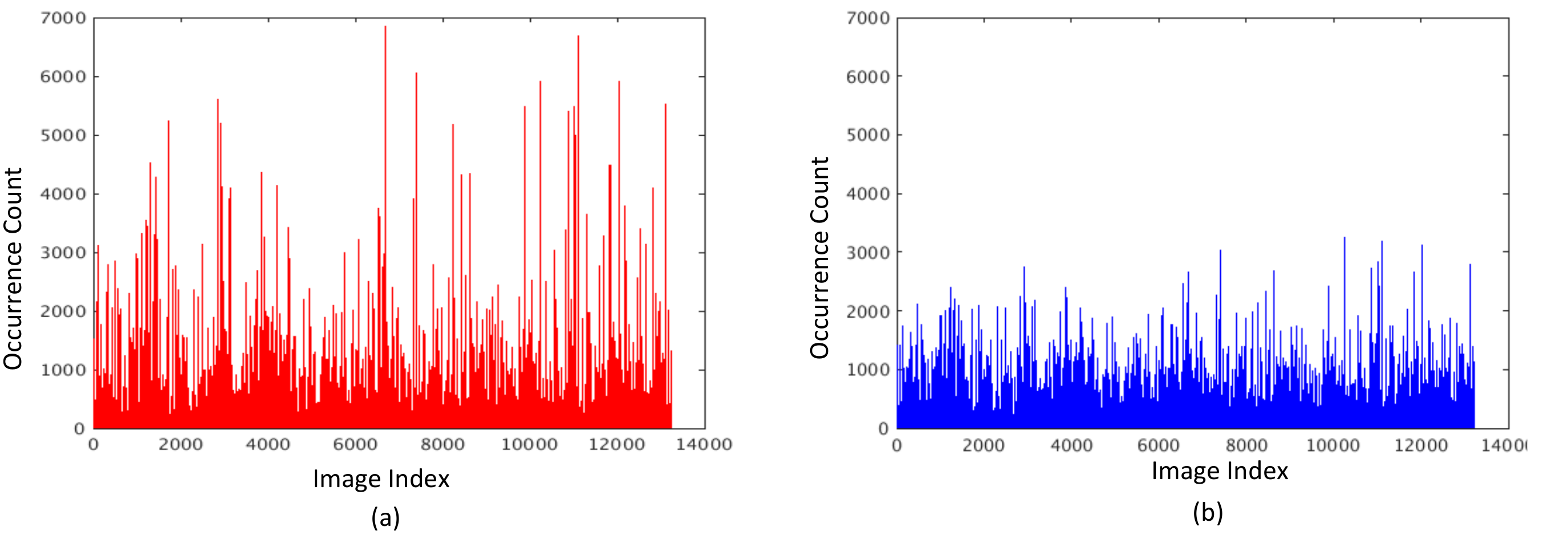}
\end{center}
    \caption{Numbers of times each face in the LFW database appeared in any other face's top-200 nearest neighbor list for a) the exact nearest neighbors, and b) the nearest neighbors computed via randomized k-d tree approximation. }
    \label{fig:term_frequency_nn}
\end{figure*}

For k-means and spectral clustering, the algorithm is parameterized on a fixed number of clusters, while for rank-order clustering the number of clusters found depends on the distance threshold parameter. In practical applications we cannot assume that the true number of clusters will be known a priori, therefore in the absence of a robust procedure for determining the true number of clusters we simply evaluate all algorithms at several effective values of $C$, and report the best results attained in Table~\ref{tab:just_lfw}.

For k-means and spectral clustering, clustering performance, per F-measure, with $C$ close to the true number of identities is quite poor (this is expected, since these algorithms are not able to handle highly unbalanced data well). For this reason, the optimal value of $C$ in terms of F-measure is relatively low (\textasciitilde200 in both cases). For rank-order clustering, the distance threshold which leads to the best overall F-measure results in a number of clusters close to the true number of identities, and the overall F-measure is significantly higher than the spectral and k-means results.

\begin{table}
\centering

\caption{Clustering Results on the LFW dataset, with approximate rank-order clustering, and LFW with additional $1$ million web-downloaded face images.  Times measured using 20 cores of an Intel Xeon CPU clocked at 2.5 GHz.}
\begin{tabular}{ l|l|r|r }
\toprule
 Nearest Neighbor Algorithm & Dataset & F-measure & Run-Time \\ \midrule
 Brute-Force & LFW & 0.72 & 00:00:12 \\
  Chen et al.~\cite{chen2009fast} & LFW & 0.69 & 00:26:36 \\
 Randomized k-d Tree~\cite{silpa2008optimised} & LFW & 0.87  & 00:00:08 \\  \midrule
 Brute-Force& LFW + 1M & 0.49 & 14:18:24 \\
  Chen et al. & LFW + 1M & 0.41 & 01:06:58 \\
  Randomized k-d Tree & LFW + 1M & 0.79  & 00:07:20 \\  \bottomrule
\end{tabular}

\label{tab:nn_lfw_1m}
\end{table}

In terms of runtime, per Table ~\ref{tab:just_lfw}, even for just $13,233$ face images in LFW spectral clustering takes noticeably large compute time, while the proposed rank-order clustering is substantially faster. Some example clusters are shown in Figure~\ref{fig:clusters_just_lfw};~\ref{fig:clusters_just_lfw}(a) and (b) show pure clusters, while ~\ref{fig:clusters_just_lfw}(c) and (d) show example impure clusters in terms of subject identity. In cluster ~\ref{fig:clusters_just_lfw}(c), $3$ images of different individuals, all with similar demographics, were grouped in with the majority identity (Walter Mondale); while in cluster ~\ref{fig:clusters_just_lfw}(d), $2$ images of $1$ additional individual with similar demographics, and face pose were grouped with the majority identity (Michael Douglas).

\subsection{Approximation Performance}

We evaluate the performance of our $k$-NN approximation method in terms of clustering accuracy, and run-time. We consider two approximation methods for computing the full $k$-NN graph, and compare their performance to the brute force approach of performing all pairwise comparisons. Results are shown in Table~\ref{tab:nn_lfw_1m} for these three nearest neighbor calculation methods on the full LFW dataset, and the LFW dataset augmented with an additional 1 million unlabeled images from the Webfaces dataset. In practice, the Randomized k-d Tree method~\cite{silpa2008optimised} achieves the best run-time of the three methods and the best clustering accuracy as well.

This is a surprising result, since an approximation method would generally be expected to give less accurate results than the process it is approximating; however, since our objective is to perform clustering based on the nearest neighbor lists, rather than simply find the exact k nearest neighbors for each item, this counter-intuitive result can be explained as follows. Figure~\ref{fig:term_frequency_nn} plots the number of times each face in the LFW dataset occurs in the top $200$ nearest neighbor list of every face in the dataset. For the exact nearest neighbors, there are a number of face images which occur very frequently in the nearest neighbor lists (up to over half of all nearest neighbor lists), while for the approximate nearest neighbors these faces occur less frequently. From the perspective of clustering based on the nearest neighbor lists, the lists computed from the randomized k-d tree approximation actually form more discriminative features, since certain faces are not present in very large fractions of the nearest neighbor lists, as is the case with the exact nearest neighbors.

\begin{table}
\centering

\caption{Clustering results using Approximate Rank-Order clustering on the LFW dataset with increasing amounts of augmented data, and different search size strategies for the approximate nearest neighbor calculations. Times measured using 5 cores for LFW+5M dataset experiments, and a single core used for the smaller experiments, on an Intel Xeon CPU clocked at 2.5 GHz.}
\begin{tabular}{ l|l|r|r}
\toprule
 Dataset & Search Size  & F-measure & Run-Time \\ \midrule
 Just LFW &  2,000  & 0.87 & 00:00:19 \\
 LFW + 1M  & 2,000 &  0.79 & 01:03:25 \\ \hline
LFW + 5M & 10,000 (linear increase) & 0.67 & 06:28:42\\
LFW + 5M & 4,000 (logarithmic increase) & 0.33 & 02:51:13\\
LFW + 5M & 2,000 (fixed) & 0.13 & 01:52:32\\ \bottomrule
\end{tabular}

\label{tab:nn_search_size_scale}
\end{table}

\begin{figure*}[t]
\begin{center}
        \includegraphics[width=0.9\textwidth,keepaspectratio]{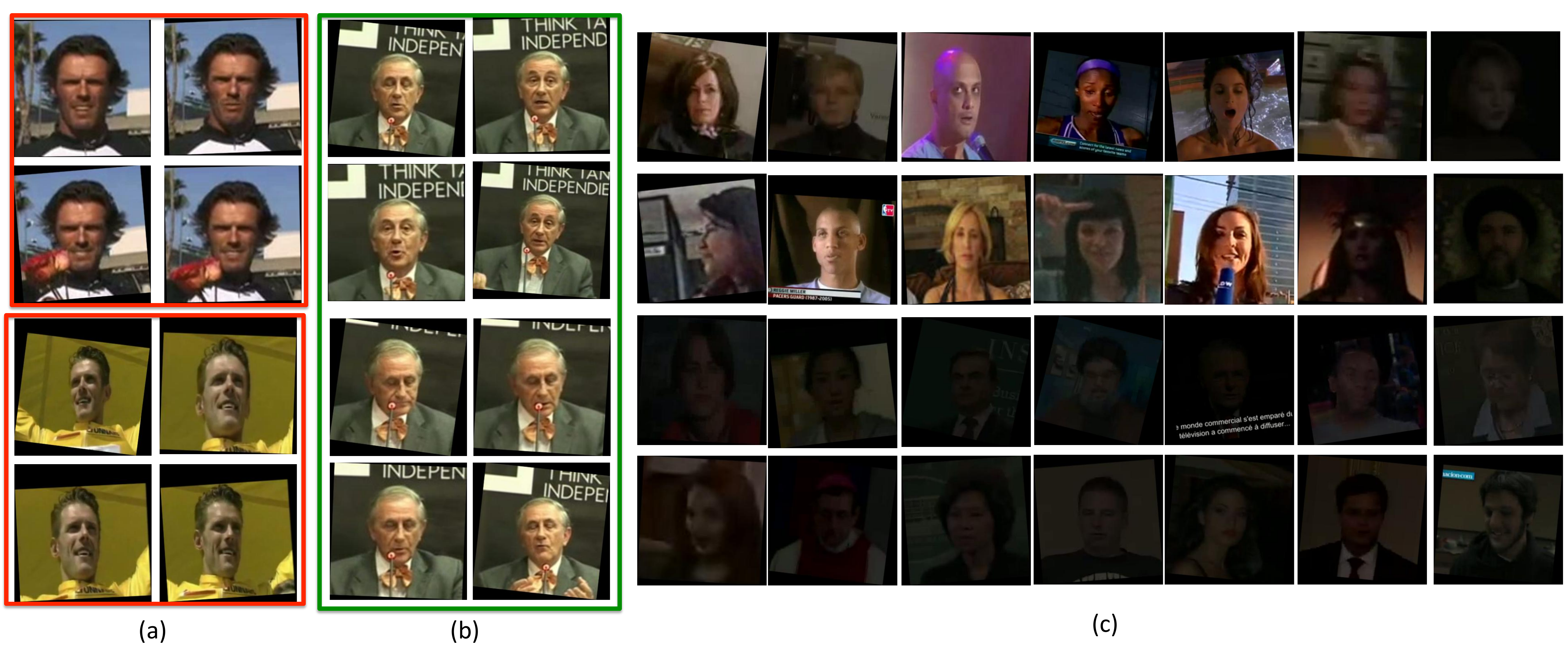}
\end{center}
    \caption{Example images from clusters generated from the YTF dataset. a) shows two clusters, each containing frames from one video of the same subject, b) shows a cluster containing frames from two videos of the same subject, where the background for the video is apparently identical, c) shows 28 identities which were incorrectly grouped into a single cluster; many of these images are poorly lit. }
    \label{fig.ytf_examples}
\end{figure*}

Generally, the randomized k-d tree algorithm has $O(n\ log\ n)$ expected run-time for tree construction, and performing $n$ searches. In practice, the FLANN implementation of the algorithm is parametrized with the number of randomized trees constructed, as well as the total number of nodes available to visit per search. If fixed parameters are used, the total runtime is indeed $O(n\ log \ n)$; however, if either the number of indices built or search size is increased with larger dataset size, the effective runtime of the algorithm will increase. In practice, we construct $4$ trees per index (and have found little impact from using slightly higher or lower values), but the number of nodes visited per search must be selected with care. One primary question is to determine if a fixed number of node visits per search is feasible for larger datasets, or if the number of nodes visited per search should increase with dataset size.  Table~\ref{tab:nn_search_size_scale} presents results for clustering based on the LFW dataset, the LFW + 1M dataset, and the LFW + 5M dataset, using different strategies for selecting the number of nodes visited per search on the LFW + 5M dataset. In practice, using the same number of nodes visited per search on the LFW + $5$M dataset as was used on the LFW + $1$M dataset leads to a drastic reduction in clustering accuracy on the larger dataset. In fact, even a logarithmic increase in search size leads to significant accuracy loss, relative to a linear increase in search size. In the following large-scale experiments, we therefore increase the search size linearly with dataset size. In practice, this means the run-time of the approximation algorithm cannot be considered to be $O(n\  log\  n)$, since we increase the cost of each search linearly with the dataset size $n$, giving a full $O(n^2)$ cost for performing $n$ nearest neighbor searches.

By using the randomized k-d tree algorithm for approximate nearest neighbor computation, with our updated clustering algorithm we get improved runtime in the clustering step, and also better clustering accuracy (compared to the baseline algorithm). Although we still have an $O(n^2)$ run-time for the nearest neighbor computation step, there is still a significant reduction in run-time, an improvement by a factor of $120$ for the LFW+$1$M image dataset over brute-force computation.

\subsection{Large-Scale Face Clustering}
In this section, we will consider clustering truly large-scale face datasets, up to $123$ million face images. As discussed, we will use the randomized k-d tree nearest neighbor approximation method to reduce the total cost of computing nearest neighbors for these datasets; however, when considering very large scale datasets, additional problems arise. Considering the total size of the dataset, $123$ million $320$-dimensional feature vectors, with each dimension represented by one float takes up approximately $157$ gigabytes of space, without considering any supporting data structures. This amount of data is difficult to fit on a single machine, considering that a tree structure must also be loaded in memory, and additionally since the approximation method we are using incurs a full $O(n^2)$ cost in time, computing nearest neighbors for this dataset becomes infeasible on a single machine.

\begin{table}
\centering

\caption{Large-scale clustering results using the proposed Approximate Rank-Order clustering, with randomized k-d tree nearest neighbor approximation. Times measured using the specified number of  cores of Intel Xeon CPUs clocked at 2.5 GHz. \# Clusters is the resulting number of clusters, excluding single-image clusters.}
\begin{tabular}{ l|r|r|r|r}
\toprule
 Dataset &  F-measure & \# Clusters & \# Cores & Run-Time \\ \midrule
 LFW  & 0.87 & 1,463 & 1 & 00:00:19 \\
 LFW + 1M  &  0.79 & 94,740 & 1 &  01:03:25 \\
LFW + 5M &  0.67 &  445,880 & 5 & 06:28:42\\
LFW + 10M & 0.56 & 933,278  & 10 & 12:11:33\\
LFW + 30M & 0.42 & 2,800,202 & 30 & 30:44:58\\
LFW + 123M &  0.27 & 10,619,853 & 123 & 289:04:53\\ \bottomrule
\end{tabular}

\label{tab:large_scale_clustering_data}
\end{table}

Fortunately, a distributed memory variation of the randomized k-d tree algorithm is available as part of the FLANN library~\cite{flann_pami_2014}. The strategy employed is to split the dataset into disjoint subsets, assign one subset to every discrete machine used, and construct separate k-d tree indices for each disjoint chunk. During nearest neighbor computation, we find a separate set of nearest neighbor candidates from each chunk, and merge the results to get a final set of nearest neighbors for the search. In practice, this simple strategy works well. In the following experiments the initial dataset is partitioned into $1$ million image chunks, and each chunk is distributed to a separate machine for index construction. Since our datasets are a small labeled subset (LFW), in a larger unlabeled background set, we randomize the order of the LFW images, and assign a portion of the labeled images to each of the discrete chunks of data, to avoid any bias due to constructing one of the sub-indices with e.g. the entire LFW dataset as part of it.

Results for progressively larger datasets (constructed by adding larger and larger sets of unlabeled background data to LFW) are presented in Table~\ref{tab:large_scale_clustering_data}. Due to our strategy of allocating 1 million images per core, we linearly increase the number of cores with dataset size, resulting in an overall $O(n)$ increase in runtime when moving to larger datasets. Computing nearest neighbors for the largest dataset considered ($123$ million images) took approximately $2$ weeks of real-time using $123$ nodes in the MSU High-Performance Computing Center. While the observed run time increases is approximately linear, the clustering accuracy progressively decays when considering larger and larger datasets. This is as expected, considering a larger dataset means a larger chance of finding impostors for each individual image as nearest neighbors. Even so, on the $123$ million image dataset, we still attain $0.27$ F-measure on the labeled subset, which is considerably better than a random result (which is close to zero, since $13,233$ images can easily be grouped into $123$M images without keeping any of the same identity face images together).

\subsection{Video Frame Clustering}

We also consider the problem of clustering video frames, using the Youtube Faces (YTF) dataset. Similar to our treatment of LFW, we cluster all faces in YTF, and evaluate the results in terms of their consistency with arranging the individual frames by identity. The results are summarized in Table~\ref{tab:ytf_clustering_data}. The overall F-measure appears reasonably consistent with our LFW results, at $0.74$ for $621,126$ total frames of video (compared to $0.79$ F-measure for clustering the LFW + 1M dataset); a lower accuracy on the YTF dataset is expected because of its generally lower image quality. However, closer analysis of the results reveals some confounding factors.

Unlike LFW clustering results, where precision and recall are relatively close for the optimal F-measure values, our clustering results on YTF have very high precision, and relatively lower recall. Effectively, we are getting more clusters than the number of identities, but the clusters are relatively pure. Further analyzing the recall, we find that although the overall value is $0.589$, the fraction of same-video pairs grouped together is much higher than the fraction of cross-video pairs grouped together. This indicates that we are successfully grouping frames into videos, but having relatively little success grouping identities across videos. Some example clusters are shown in Figure~\ref{fig.ytf_examples}. In most cases, clusters roughly correspond to single videos, in a few cases, e.g. ~\ref{fig.ytf_examples}(b), frames from different videos of the same individual are correctly grouped together, and in a small subset of clusters (e.g. ~\ref{fig.ytf_examples}(c)), multiple identities are grouped in the same cluster.

These results indicate some weaknesses in our clustering algorithm. One confounding factor is the nature of the fixed nearest neighbor lists we use. We generate the top $200$ nearest neighbors for face each image, but in a video dataset, it is possible, and even likely that all $200$ of the closest neighbors are other frames of the same video (since a subjects appearance within a video will typically change less than across videos). One potential strategy would be to first conduct clustering, attempting to consolidate the different identities present within a video (this is not possible to evaluate using the YTF database, since only one identity is labeled per video in YTF), then perform clustering again on a reduced dataset to consolidate the per-video identities across different videos.

\begin{table}
\centering

\caption{YTF clustering results using the proposed Approximate Rank-Order clustering, with randomized k-d tree nearest neighbor approximation. Time is given as HH:MM:SS, measured using 20 cores of a Intel Xeon CPU clocked at 2.5 GHz.}
\begin{tabular}{ l|r}
\toprule
 Performance Measure &  Value \\ \midrule
 F-Measure  & 0.71\\ \midrule
Precision  &  0.79 \\
Recall &  0.67 \\ \midrule
Within-Video Recall & 0.56 \\
Cross-Video Recall & 0.91 \\ \midrule
Nearest Neighbor Computation Time & 00:04:10\\ 
\bottomrule
\end{tabular}

\label{tab:ytf_clustering_data}
\end{table}

\subsection{Per-Cluster Quality: Internal Measures}\label{sec:internal}

We are interested in identifying a good subset of clusters from a large group of clusters, as a means of aiding manual exploration of large datasets. To evaluate the effectiveness of different measures experimentally, we need methods for evaluating the effectiveness of the internal measures. As a first approach, we consider the correlation between the internal measures and an external measure, the pairwise precision computed individually for each cluster. Correlation between the various internal measures considered, and precision are show in Table~\ref{tab:internal_external_correlations}. In practice, the graphical measures do not perform particularly well, while the simpler measures based on edge weights perform better. In particular, the best correlation is observed for the ``average inter-cluster edge weight", and this can be further improved by subtracting the average inter and intra cluster edge weights. This is reasonable, since we are effectively combining a compactness measure (average intra cluster edge weight), and a separability measure (average inter-cluster edge weight). Even so, the best correlation achieved is only $0.42$. This correlation can be improved by excluding size $2$ clusters from consideration (so only examining clusters with $3$ or more members), which improves correlation to $0.46$. 

\begin{table}
\caption{Correlation of internal cluster quality measures with precision, for the LFW dataset. Average precision@100 is the unweighted average of per-cluster pairwise precision for the top-100 scoring clusters for each metric.}
\begin{tabularx}{\columnwidth}{X|r|r}
\toprule
 Internal Measure &  Correlation & Avg. Precision@100  \\ \midrule
Inter-MQ  & 0.128  & 0.748\\
Intra-MQ  &  0.117 & 0.837\\
MQ (Combined) &  0.120 & 0.829 \\ 
Coverage & 0.117 & 0.748\\
\hline
Max Intra-cluster Edge Distance & 0.022 & 0.860\\
Total Intra-Cluster Edge Distance & 0.030 & 0.850\\ 
Average Intra-Cluster Edge Distance & 0.080 & 0.853\\
\hline
Minimum Inter-Cluster Edge Distance & 0.205 & 0.893\\
Total Inter-Cluster Edge Distance & 0.125 & 0.852 \\
Average Inter-Cluster Edge Distance & 0.325 & 0.880\\
\hline
(Avg. Intra-Cluster Edge Distance) - (Avg. Inter-Cluster Edge Distance) & 0.427 & 0.908\\
(Avg. Intra-Cluster Edge Distance) - (Avg. Inter-Cluster Edge Distance, cluster size $\ge$ 3) & 0.460 &0.979\\
\bottomrule
\end{tabularx}

\label{tab:internal_external_correlations}
\end{table}

\begin{figure}[t]
  \centering
    \includegraphics[width=\columnwidth]{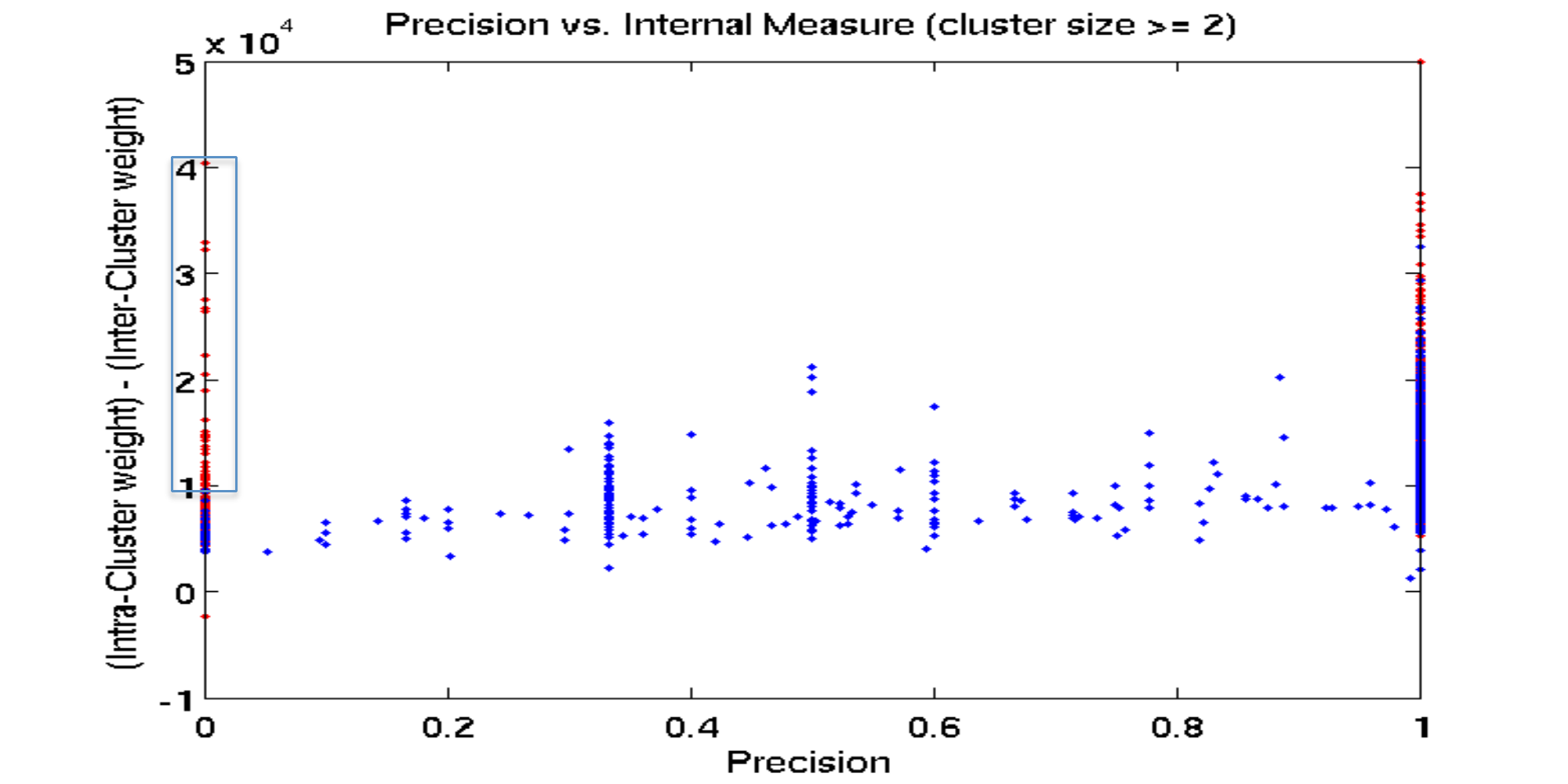}
  \caption{Pairwise precision vs. the proposed internal quality measure, for all clusters generated by the proposed Approximate Rank-Order clustering algorithm on the full LFW dataset. Points in blue are clusters of size $3$ or larger, points in red are of size $2$. The highlighted set of points on the left edge of the figure are all of size $2$, with zero pairwise precision. Since we can't reliably distinguish between good and bad $2$-item clusters, we discard them from consideration. }
  \label{fig:lfw_internal_external_src}
\end{figure}

While a correlation of $0.46$ is not very high, for our application there is no particular need for the relationship between the internal and external measures to be linear. Figure~\ref{fig:lfw_internal_external_src} plots the external measure (Precision) vs. the best performing internal measure \{(avg. intra-cluster edge weight) - (avg. inter-cluster edge weight)\}. One notable feature on the left side of the plot is a set of clusters with exactly zero precision, that still score relatively highly on the internal measure. Closer examination reveals that all of these high scoring zero-precision clusters are of size two (so they consist of relatively isolated faces of different people that happen to score highly), which explains the improvement in correlation (from $0.42$ to $0.46$) when restricting consideration to size 3 or larger clusters. Another interesting feature of Fig.~\ref{fig:lfw_internal_external_src} is that the points on the plot almost form a triangle (with a variety of precision values for low-scoring clusters, but mostly just high precision values for high-scoring clusters), so although the relationship between the external and internal measures is not linear, it is still possible to select a subset of high precision clusters by taking a high threshold on the internal measure.

\subsubsection{Ranking Evaluation}

\begin{figure}[t]
  \centering
    \includegraphics[width=\columnwidth]{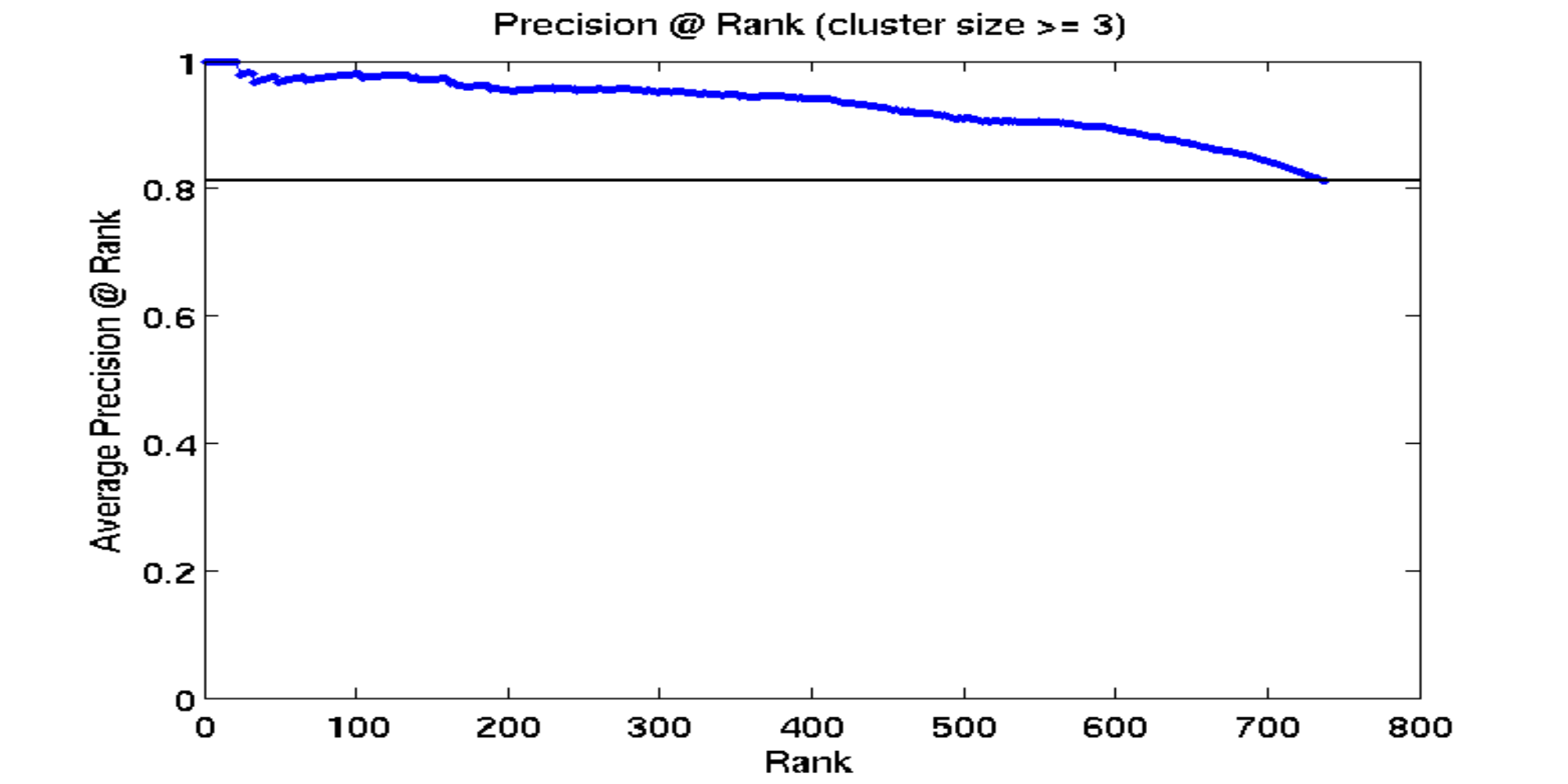}
  \caption{Average pairwise precision for lists of clusters ordered by the proposed internal cluster quality measure, terminated at each possible rank. The horizontal black line indicates the unweighted average precision of all clusters considered. Clusters are generated by the proposed Approximate Rank-Order clustering algorithm from the full LFW dataset.}
  \label{fig:lfw_precision_at_rank}
\end{figure}

\begin{figure*}[t]
        \includegraphics[width=\textwidth]{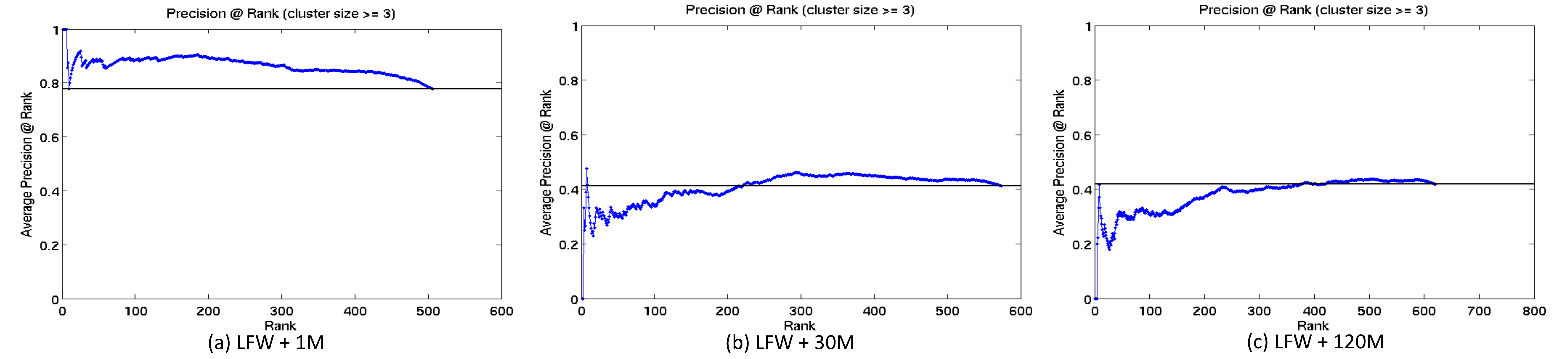}
    \caption{Average pairwise precision at rank, ordered by the proposed internal cluster quality measure for augmented datasets. Clusters are generated by the proposed Approximate Rank-Order clustering algorithm.}
    \label{fig:precision_at_rank_multi}
\end{figure*}

As an alternative to considering correlation, we can use the internal measure to rank all the clusters, and compute the unweighted average of per-cluster precision values for the top $C$ clusters (ranked by the internal measure), inspired by analysis typically done in retrieval problems. Table~\ref{tab:internal_external_correlations} shows the average precision of the various internal measures considered for the top-100 clusters in the full LFW dataset. Figure~\ref{fig:lfw_precision_at_rank} shows a plot of the average precision of the top $C$ clusters, with $C$ cut off at each possible rank in the sorted list of clusters, for the best performing internal measure.  The internal measure is effective in selecting high precision clusters (relative to the average precision of all the clusters) on the LFW dataset.  In fact, the first several clusters ranked by the internal measure have a pairwise precision of 1. Figure~\ref{fig:precision_at_rank_multi} extends this concept to the augmented datasets (in this case, only clusters containing some labeled data from LFW are ranked, and precision is computed omitting unlabeled clusters as in our previous evaluation).

\begin{figure*}[t]
        \includegraphics[width=\textwidth]{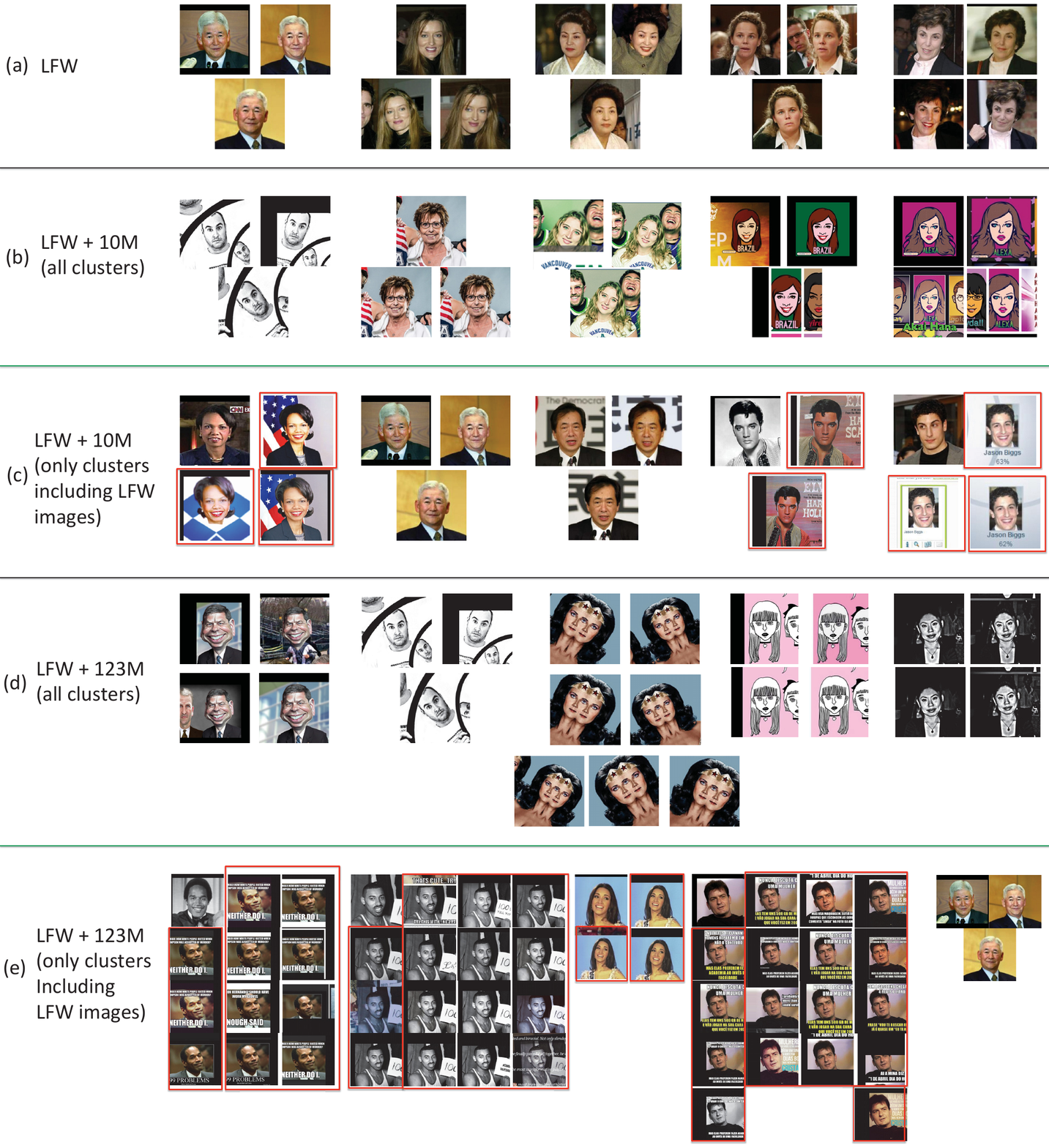}
    \caption{Top-5 ranked clusters for the LFW, LFW+10M, and LFW+123M datasets. For the LFW+10M, and LFW+123M datasets, both the absolute top-5 ranking clusters in terms of the proposed quality measure, and the top-5 ranking clusters out of those clusters containing at least some LFW images are shown. Unlabeled background images grouped in with the LFW images in the LFW+10M and LFW+123M datasets are outlined in red, these images are all of the same subject as the LFW images in each cluster, showing the strength of the proposed quality measure.}
    \label{fig:ranking_examples}
\end{figure*}

Initially, the internal measure is still effective; however, for very large datasets (LFW+ $30$M and above), ranking clusters according to the internal measure, as expected, becomes less effective. Some example top ranking clusters are shown in Figure~\ref{fig:ranking_examples}.  The top-$5$ clusters for the LFW dataset are all single identity, relatively small clusters, indicating that the quality measure works as expected. For larger datasets (LFW+$10$M, LFW+$123$M), we show both the top-$5$ clusters ranked purely in terms of the quality measure, (b) and (d), as well as the top-$5$ results containing any labeled data, (c) and (e) (since we use the labeled subset in our numerical evaluations). The top clusters in absolute ranking typically involve near-duplicate images (e.g. similar images uploaded in different locations, with minor differences due to cropping, resolution, or color correction differences), and often cartoon faces (which were detected by face detectors) in addition to actual photographs.

The top clusters involving LFW images show that there are in fact a number of images of the LFW subjects in the unlabeled dataset--this indicates that our performance evaluation is to a certain extent overly conservative, since we consider grouping LFW and unlabeled data together to be incorrect (due to lack of label information). Although the results for the LFW+$10$M dataset appear reasonable (in the sense that multiple images of the same identity are being grouped together, that are not just slight alterations of the same original image), for the LFW+$123$M dataset we begin to see a large number of near duplicate images, e.g. the clusters ranked $1$ through $4$ on the list of clusters with LFW images have a single LFW image, and multiple near duplicate images that happened to be in the background set. Nevertheless, these clusters appear to be pure, and the proposed clustering algorithm is meeting its objective.

\section{Conclusions}

We have shown the feasibility of clustering a large collection of unlabeled face images (up to $123$M) into an unspecified number of identities (on the order of millions). This problem is of practical interest as a first step in organizing a large collection of unlabeled face images prior to human examination due to the high volume of face images uploaded to social media, and potentially encountered in forensic investigations. There are computational challenges in processing datasets with tens of millions of faces (which we address via approximation methods, and parallelization). Even if the computational challenges are met, producing meaningful clusters on data of this scale is very difficult. In terms of clustering accuracy, we achieved $0.27$ pairwise F-measure on the largest dataset considered ($123$M unlabeled faces + $13,233$ labeled images from LFW), which indicates that at least some of the clusters produced by our algorithm correspond well to true identities in LFW. To identify these high quality clusters, we developed an internal per-cluster quality measure, that does not involve external identity labels, to rank the clusters by quality for manual examination. Experimental results showed that this measure was extremely effective for smaller datasets, but for the larger datasets considered (LFW + $123$M unlabeled faces), performance, as expects falls.  Still, some good quality (compact and isolated) face clusters can be identified. 

In terms of future work, while the underlying face representation we employ works reasonably well for unconstrained face images, it could still be improved in a number of ways (e.g. using larger training sets, or improving the deep model architecture). While we were able to apply our clustering algorithm to datasets up to $123$ million face images, we need to improve the clustering method (e.g. by incorporating more accurate nearest neighbor methods) to obtain better clustering accuracy. Other areas for improvement include the automatic selection of the number of clusters in a fully unlabeled dataset, as well as improving our per-cluster quality evaluation methods, and utilizing pair-wise constraints (must-link and cannot-link) to improve clustering accuracy. 


\section*{Acknowledgements}
We would like to thank the Noblis corporation for their assistance in acquiring the unlabeled background images used in this work.

\bibliographystyle{IEEEtran}
\bibliography{IEEEabrv,face_clustering_journal}


\newcommand{\biospace}{\vspace{-0.5in}}
\biospace
\begin{IEEEbiography}[{\includegraphics[width=1in,height=1.25in,clip,keepaspectratio]{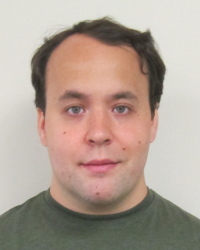}}]{Charles~Otto} received his B.S. degree in the Department
of Computer Science and Engineering at Michigan State University in 2008. He was a research engineer at IBM during 2006-2011. Since 2012, he has been working towards the Ph.D. degree
in the Department of Computer Science and Engineering at Michigan State University. His research interests include pattern recognition, image processing, and computer vision, with applications to face recognition.
\end{IEEEbiography}

\biospace
\begin{IEEEbiography}[{\includegraphics[width=1in,height=1.25in,clip,keepaspectratio]{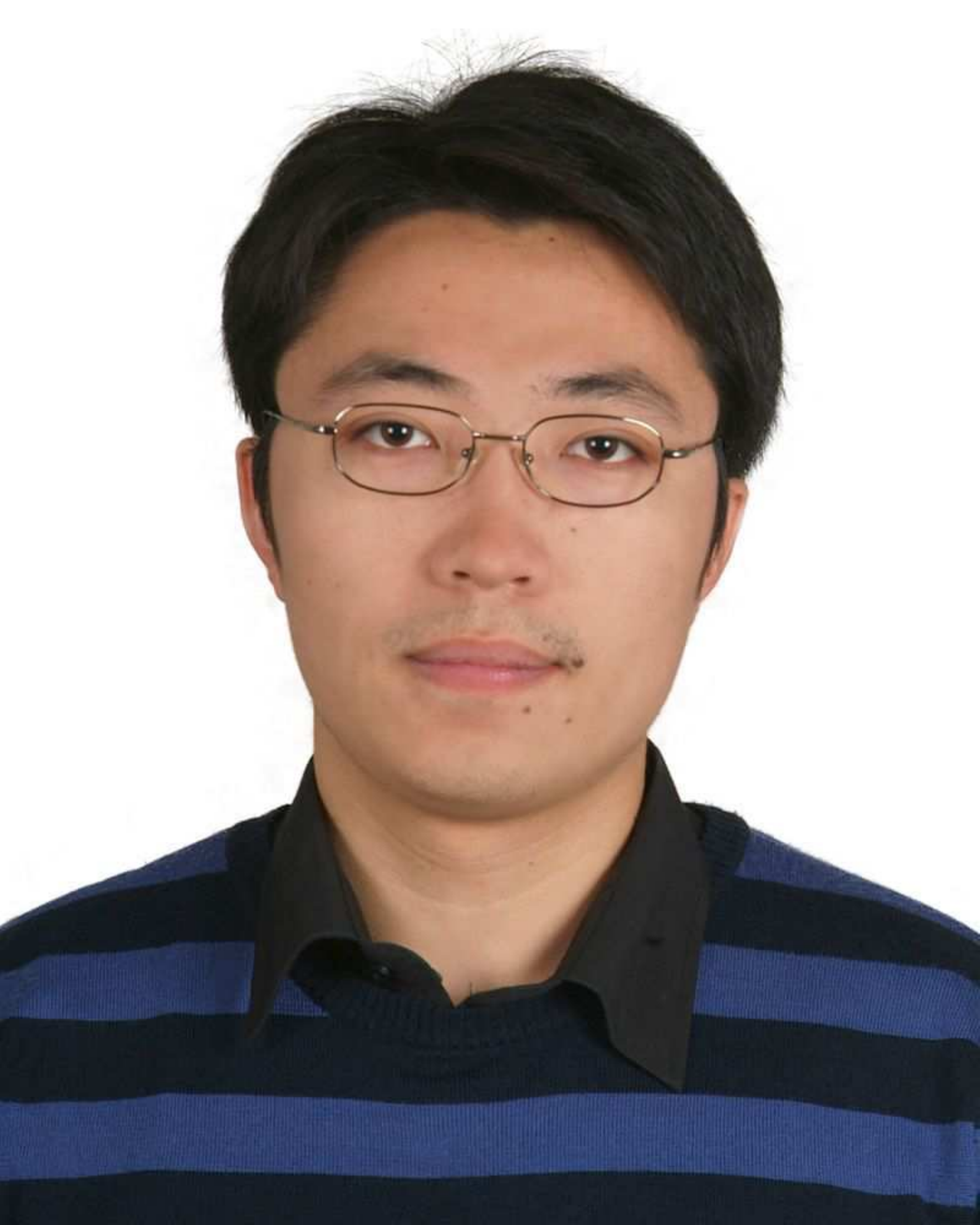}}]{Dayong~Wang} received his bachelor degree from Tsinghua University in 2008 and his Ph.D. degree from Nanyang Technological University, Singapore, 2014. He is currently a Postdoctoral Researcher at Michigan State University, USA. His research interests are statistical machine learning, pattern recognition, and multimedia information retrieval. In his research areas, he has published several papers in top venues, including TPAMI, TKDE, ACM MM, and SIGIR.
\end{IEEEbiography}
\biospace
\begin{IEEEbiography}[{\includegraphics[width=1in,height=1.25in,clip,keepaspectratio]{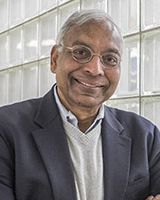}}]{Anil~K.~Jain} is a University distinguished professor in the Department of Computer Science and Engineering at Michigan State University. His research interests include pattern recognition and biometric authentication. He served as the editor-in-chief of the IEEE Transactions on Pattern Analysis and Machine Intelligence (1991-1994), a member of the United States Defense Science Board, and The National Academies committees on Whither Biometrics and Improvised Explosive Devices. He has received Fulbright, Guggenheim, Alexander von Humboldt, and IAPR King Sun Fu awards. He was elected to the National Academy of Engineering in 2016.
\end{IEEEbiography}
\end{document}